\documentclass[letterpaper, 10 pt, conference]{ieeeconf}
\usepackage[binary-units=true]{siunitx}
\usepackage{times}
\usepackage{amsmath,amssymb}
\usepackage[nolist,nohyperlinks]{acronym}
\usepackage{accents}
\usepackage{graphicx}
\usepackage{bm}
\usepackage{bbm}
\let\labelindent\relax
\usepackage[inline]{enumitem}
\usepackage{array}
\usepackage{multicol}
\usepackage{multirow}
\usepackage{tabulary}
\usepackage{booktabs}
\usepackage{subfig}
\usepackage{todonotes}
\usepackage{cite}
\usepackage{algorithm}
\usepackage{algpseudocode}
\usepackage[bookmarks=true]{hyperref}
\usepackage{xcolor}

\newcommand{\norm}[1]{\left\lVert#1\right\rVert}
\DeclareMathOperator*{\argmax}{arg\,max}
\DeclareMathOperator*{\argmin}{arg\,min}
\begin{acronym}
	\acro{CoM}{Center of Mass}
	\acro{DFS}{Depth First Search}
	\acro{CITO}{Contact-Implicit Trajectory Optimization}
	\acro{MIP}{Mixed-Integer Programming}
	\acro{MIQP}{Mixed-Integer Quadratic Programming}
	\acro{RNN}{Recurrent Neural Network}
	\acro{MLP}{Multilayer Perceptron}
	\acro{S}[\emph{S}]{\textbf{sliding}}
	\acro{SC}[\emph{SC}]{\textbf{sliding with curvature}}
	\acro{R}[\emph{R}]{\textbf{rotating}}
	\acro{P}[\emph{P}]{\textbf{pivoting}}
	\acro{L}[\emph{L}]{\textbf{lifting}}
	\acro{MCTS}{Monte Carlo Tree Search}
	\acro{PVMCTS}{Policy-Value Monte Carlo Tree Search}
	\acro{MDP}{Markov-decision Process}
	\acrodefplural{MDP}{Markov-decision Processes}
	\acro{ADMM}{Alternating Direction Method of Multipliers}
	\acro{QP}{Quadratic Program}
	\acro{DoF}{Degree of Freedom}
	\acrodefplural{DoF}{Degrees of Freedom}%
	\acro{leg}[\emph{L}]{\emph{legible}}		
\end{acronym}
\IEEEoverridecommandlockouts          
\renewcommand{\baselinestretch}{0.98}

\title{\LARGE \bf
Efficient Object Manipulation Planning with Monte Carlo Tree Search}

\author{Huaijiang Zhu$^{1}$, Avadesh Meduri$^{1}$, Ludovic Righetti$^{1}$%
\thanks{$^{1}$Tandon School of Engineering, New York University, USA}%
\thanks{This work was in part supported by the National Science Foundation (grants 1932187, 1925079, 2026479 and 2222815) and Meta Platforms, Inc.}
}

\begin{document}
\maketitle
\thispagestyle{empty}
\pagestyle{empty}
\begin{abstract}
    This paper presents an efficient approach to object manipulation planning using Monte Carlo Tree Search (MCTS) to find contact sequences and an efficient ADMM-based trajectory optimization algorithm to evaluate the dynamic feasibility of candidate contact sequences.
    To accelerate MCTS, we propose a methodology to learn a goal-conditioned policy-value network and a feasibility classifier to direct the search towards promising nodes. Further, manipulation-specific heuristics enable to drastically reduce the search space. Systematic object manipulation
    experiments in a physics simulator and on real hardware demonstrate the efficiency of our approach. In particular, our approach scales favorably for long manipulation sequences thanks to the learned policy-value network, significantly improving planning success rate. All source code including the baseline can be found at \url{https://hzhu.io/contact-mcts}.
\end{abstract}

\section{Introduction}
The ability to plan sequences of contacts and movements to manipulate objects 
is central to endow robots with sufficient autonomy to perform complex tasks.
This remains, however, particularly challenging. Indeed, finding dynamically
feasible sequences of contacts between the manipulator and an object typically leads
to intractable combinatorial and nonlinear problems. Consider a simple task of reorienting an object resting on a table with a two-fingered hand. To plan suitable contacts, it is necessary to reason about interaction forces. For example, a cube can be rotated by applying forces from the sides; however, if the object is a thin plate, the fingers must ``press down'' the object to create friction to achieve the same task. More importantly, as the fingers reach their respective kinematic limits, they need to break and re-establish contacts to rotate the object further. These two aspects, interaction forces and contact switches, have been the main challenges of object manipulation planning.

Over the past decade, trajectory optimization has been favored for multi-contact motion planning as this leads to efficient formulations to reason about interaction forces~\cite{Escande_Kheddar_2009,Lin_Righetti_Berenson_2020,Ponton_Khadiv_Meduri_Righetti_2021,Carpentier_Tonneau_Naveau_Stasse_Mansard}. Yet, it remains unclear how the planning of contact modes should be efficiently incorporated, primarily due to its discrete nature that results in discontinuity in the dynamics at contact switch. To handle this discontinuity under the trajectory optimization framework, two main streams of methodologies have emerged:
\begin{enumerate}
    \item the contact-invariant or contact-implicit approach enforces contact complementarity either as hard constraints~\cite{stewart2000implicit, posa2014direct},  penalty terms in the cost function~\cite{mordatch2012discovery, mordatch2013animating, mordatch2012contact}, or with differentiable contact models~\cite{neunert2018whole, pang2022global}, and
    \item the hybrid approach treats contact switches as discrete decisions and incorporates them in the continuous trajectory optimization problem~\cite{toussaint2015logic, aceituno2020global, doshi2020hybrid, cheng2022contact, chen2021trajectotree}.
\end{enumerate}

In the context of robot manipulation, one representative work is Contact-Trajectory Optimization proposed in~\cite{aceituno2020global}, where contact scheduling is modeled as binary decision variables and the non-convexity of the dynamics due to cross products is relaxed using McCormick envelopes~\cite{mccormick1976computability}. The problem can then be formulated as \ac{MIQP}~\cite{lazimy1982mixed}. However, the usage of the McCormick envelopes leads to a relaxed problem instead of the original one. As a result, applying the planned contact forces and locations may incur undesired torques. To date, the approach has only been demonstrated on 2D object manipulation with very short manipulation sequences, probably because its complexity grows exponentially with respect to the number of discrete variables due to the mixed-integer formulation.

Recent research has made progress towards speeding up \ac{MIP} ~\cite{floudas1995nonlinear} by leveraging machine learning techniques. For example, Nair et al. use neural networks to learn branch-and-bound heuristics and partial assignment for the discrete variables~\cite{nair2020solving}; CoCo~\cite{cauligi2021coco} finds feasible solution to \ac{MIP} by learning to assign discrete variables and solving the resulted convex optimization problem. While such methodology greatly improves the solution speed at inference time, it assumes that the original \ac{MIP} can be solved in a reasonable amount of time to construct the training set. If the original problem is prohibitive to solve, collecting a large dataset for this problem may not be practical without abundant computational resources or additional learning of a problem reduction~\cite{lin2022reduce}.

In this work, we approach the problem from a different angle. Instead of modeling and solving it as a \ac{MIP}, we formulate a tree search problem to find dynamically consistent contact sequences. Specifically, we use learning-based \ac{MCTS}~\cite{silver2017mastering} which has recently been shown to outperform \ac{MIQP} in gait discovery~\cite{amatucci2022monte}; then we formulate the underlying continuous optimization problem as a Biconvex Program~\cite{gorski2007biconvex} to allow efficient solution via the \ac{ADMM}~\cite{boyd2011distributed} which has been adopted in online whole-body motion planning due to its guaranteed sublinear rate of convergence~\cite{meduri2023biconmp}. This leads to a formulation in which the discrete search space can be significantly reduced by introducing domain-specific heuristics for robot manipulation and the continuous problem can be solved efficiently without relaxation. More importantly, we show that learning-based \ac{MCTS} trained on short-horizon tasks generalizes directly to long-horizon tasks. This removes the need for collecting data on large-scale problems, which can be time consuming. 
The contributions of the paper are as follows:
\begin{enumerate}
    \item formulation of dynamically consistent contact planning for manipulation using learning-based \ac{MCTS},
    \item efficient resolution of the underlying continuous optimization problem as a Biconvex Program with \ac{ADMM}, and
    \item extensive simulation including comparisons with \ac{MIQP} approaches as well as real robot experiments to demonstrate the capabilities of our approach.
\end{enumerate}
To our best knowledge, this is the first application of learning-based \ac{MCTS} to dynamically consistent contact planning for manipulation.
\section{Problem Statement}
\label{sec:prob}
\subsection{Inputs}
We aim to solve an object manipulation task similar to the Contact-Trajectory Optimization problem proposed in~\cite{aceituno2020global} where the following quantities are given:
\begin{enumerate}
    \item a rigid object with known geometry, friction coefficient $\mu$, mass $m$, moment of inertia $\mathcal{I}$, and $N_{\Omega}$ pre-defined contact surfaces, each of which can be described as the convex span of its vertices $v_{\Omega, i}$,
    \item a trajectory with discretization step $\Delta t$ of length $T$ that consists of the desired object pose, velocity, and acceleration $\xi = [q(t), \dot{q}(t), \ddot{q}(t)]_{t=1}^{T}$, where $q(t) = [p(t), R(t)] \in \mathrm{SE}(3)$ consists of the position and orientation, $\dot{q}(t) = [v(t), \omega(t)] \in \mathfrak{se}(3)$ consists of the linear and angular velocity, and $\ddot{q}(t) = [\dot{v}(t), \dot{\omega}(t)]$ is the acceleration,
    \item an environment with known geometry and friction coefficient $\mu_e$, and
    \item a robot with $N_c$ end-effectors that are able to make point contact with the object.
\end{enumerate}

At the $t$-th time step, given the object motion and the object dynamics, we can compute the desired total force $f_{\text{des}}(t)$ and torque $\tau_{\text{des}}(t)$ to be applied to the object
\begin{align}
\begin{bmatrix}
f_{\text{des}}(t)\\
\tau_{\text{des}}(t)
\end{bmatrix} = 
\begin{bmatrix}
m(\dot{v}(t) + \omega(t) \times v(t) - g(t))\\
\mathcal{I}\dot{\omega}(t)+\omega(t)\times\mathcal{I}\omega(t)
\end{bmatrix}\,,
\end{align}
where all quantities, including the gravity term $g(t)$ are expressed in the \textbf{object frame.}

In addition, as the geometry of the object and the environment as well as the object motion are known, we can obtain $N_e(t)$ environment contact locations $r_e(t)$ for $e \in \{N_c + 1,\dots,N_c + N_e(t)\}$ at each time step $t$ by checking the collisions between the object and the environment, assuming uniform pressure distribution.
\subsection{Outputs}
\begin{figure}
    \centering
    \includegraphics[width=.35\textwidth]{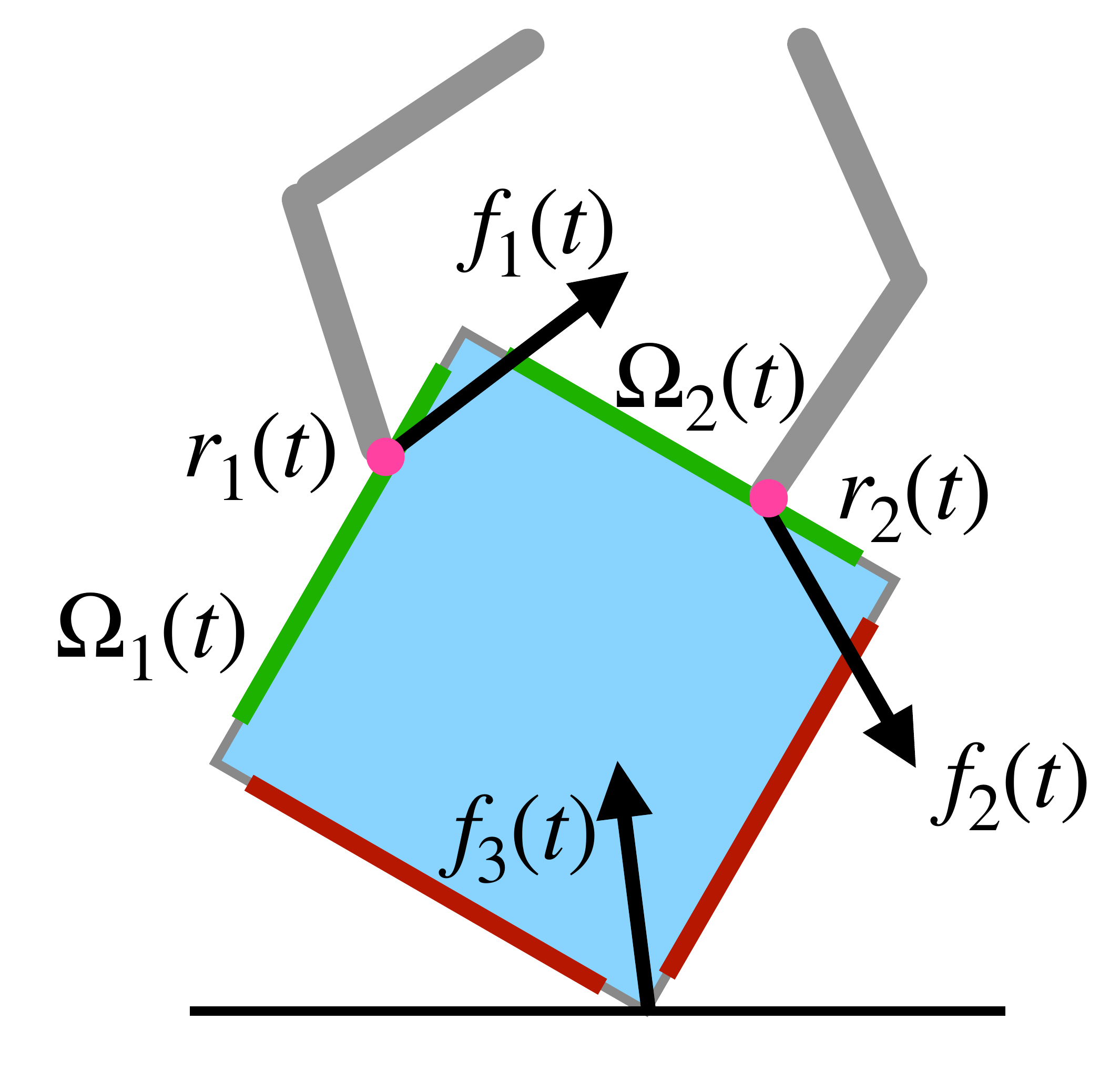}
    \caption{Illustration of the outputs of the method. $f_1(t),f_2(t)$ are the manipulator contact forces and $r_1(t), r_2(t)$ are the respective contact locations (marked by the pink dots.) $f_3(t)$ is the environment reaction force while the environment contact location is known beforehand. The thick lines on the object depict pre-defined contact surfaces and the green ones are the selected contact surfaces $\Omega_1(t), \Omega_2(t)$ at time $t$.}
    \label{fig:outputs}
    \vspace{-5mm}
\end{figure}

We aim to find the following:
\begin{enumerate}
    \item the contact surface $\Omega_c(t) \in \{0,1,\dots,N_{\Omega}\}$, the contact force $f_c(t)$ and the contact location $r_c(t)$ for each end-effector $c$ of the robot; $\Omega_c(t) = 0$ indicates that the $c$-th end-effector is not in contact, and
    \item the environment contact force $f_e(t)$
\end{enumerate}
such that the forces and torques sum to the desired ones. Fig.~\ref{fig:outputs} illustrates the outputs of our method for a double-finger manipulator pivoting a 2D square. Note that this is only for illustrative purpose and our method is applicable to and tested on 3D objects and $\mathrm{SE}(3)$ poses as we will show in the experiments.

\section{Method}

\begin{figure}[t]
    \centering
    \includegraphics[width=.45\textwidth]{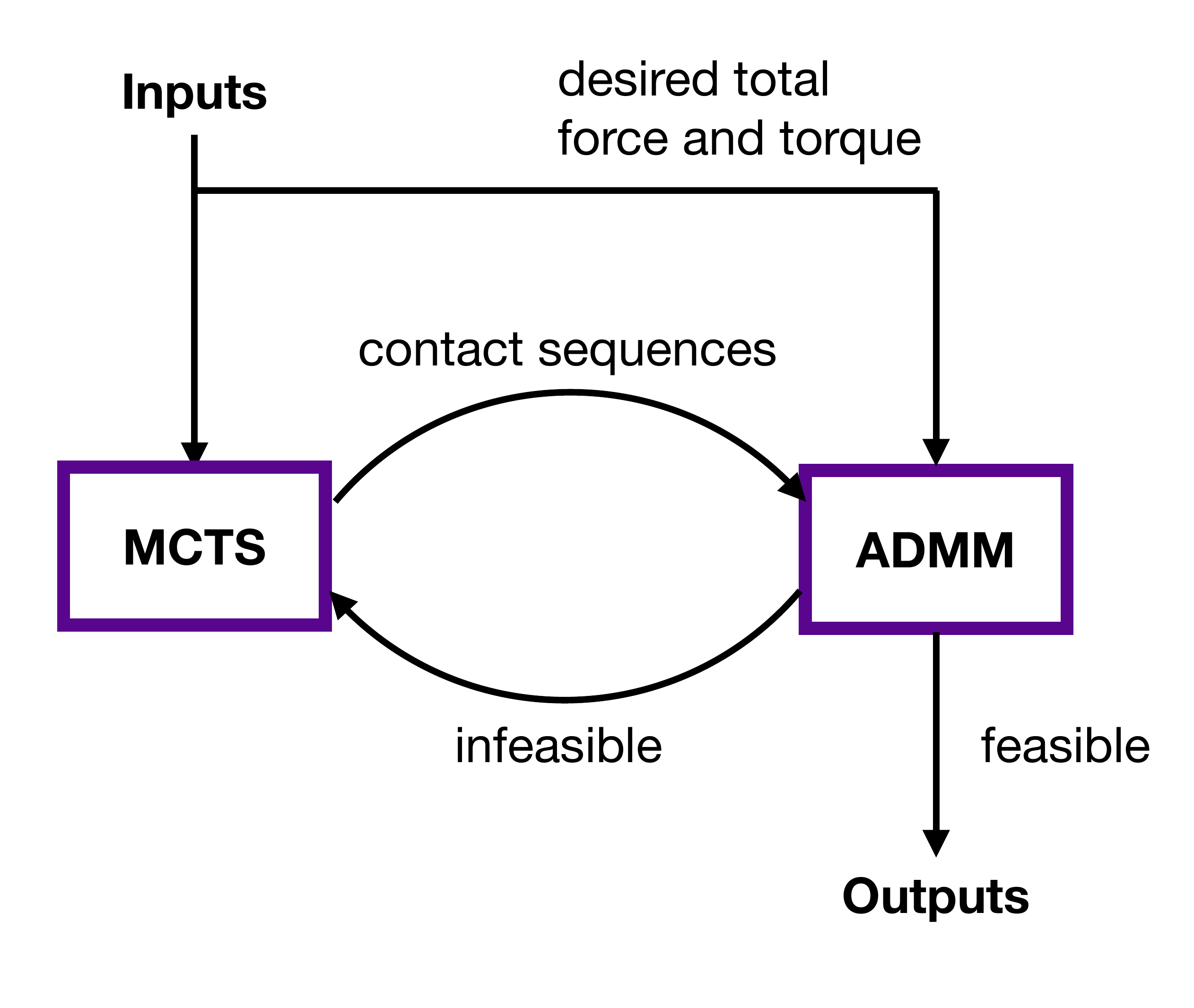}
    \caption{An overview of the proposed method. First, candidate contact sequences are proposed by MCTS. Then, they are evaluated by an ADMM-based trajectory optimization algorithm to find dynamically feasible contact forces and locations to realize the desired object motion. At inference time, this repeats until the first feasible solution is found; at training time, we let the algorithm discover multiple solutions and collect both feasible and infeasible contact sequences to construct a diverse training set.}
    \label{fig:overview}
\end{figure}

The problem described above is difficult even though the desired object motion is provided, as the solver needs to decide not only the contact force and location, but also the timing of contact switches. To address this challenge, we use learning-based \ac{MCTS} to discover promising contact sequences and then evaluate their dynamical feasibility using an \ac{ADMM}-based trajectory optimization algorithm. Fig.~\ref{fig:overview} provides an overview of our approach.

\subsection{Continuous Contact Optimization via ADMM}
\label{sec:cont_opt}
First, let us consider a simpler sub-problem where we already obtained a sequence of contact surfaces $[\Omega_1(t),\dots,\Omega_{N_c}(t)]_{t=1}^{T}$ for each end-effector $c$. We can find the exact contact forces and locations by solving a non-convex continuous optimization problem. However, this non-convex problem, as we will show, can be formulated as a Biconvex Program and solved efficiently with \ac{ADMM}. In contrast to~\cite{aceituno2020global}, our formulation does not require piecewise linear approximation of the cross product and solves the exact original non-convex problem instead of a relaxed one. The optimization problem can be described by the following constraints and cost function:

\subsubsection{Dynamics}
The contact forces and torques must sum to the desired ones
\begin{subequations}
\label{eq:dyn_constr}
\begin{align}
    \sum_{c=1}^{N_c} f_c(t) + \sum_{e=N_c + 1}^{N_c + N_e(t)} f_e(t) &= f_{\text{des}}(t)\label{eq:newton_constr}\\
    \sum_{c=1}^{N_c} r_c(t) \times f_c(t) + \sum_{e=N_c + 1}^{N_c + N_e(t)} r_e(t) \times f_e(t) &= \tau_{\text{des}}(t)\,.\label{eq:euler_constr}
\end{align}
\end{subequations}

\subsubsection{Contact location}
The contact location must be inside the given contact surface $\Omega_c(t)$ for $\Omega_c(t)\neq 0$
\begin{subequations}
\label{eq:contact_location}
\begin{align}
\forall c\in\{c|\Omega_c(t)\neq 0\}\,, &\sum_{i=1}^{N_{v, \Omega_c(t)}} \alpha_{c,i}(t) v_{\Omega_c(t),i} = r_c(t)\,,\\
&\sum_{i=1}^{N_{v, \Omega_c(t)}} \alpha_{c,i}(t) = 1\,,\\
&\sum_{i=1}^{N_{v, \Omega_c(t)}} \alpha_{c,i}(t) \geq 0\,,
\end{align}
\end{subequations}
where $v_{\Omega_c(t), i}$ is the $i$-th vertex and $N_{v, \Omega_c(t)}$ the number of vertices of the contact surface $\Omega_c(t)$ for the end-effector $c$.

\subsubsection{Contact force}
If the $c$-th end-effector is not in contact with any contact surface, hence $\Omega_c(t) = 0$, the contact force is set to zero
\begin{align}
\label{eq:force_complementarity}
\forall c\in\{c|\Omega_c(t) = 0\}\,, f_c(t) = 0\,.
\end{align}
Note that this is not a complementarity constraint as $\Omega_c(t)$ is already given by \ac{MCTS}.

\subsubsection{Sticking contact}
To prevent the end-effector from sliding on the object, we impose that if the end-effector is in contact at time step $t$, it must remain sticking at time step $t + 1$
\begin{align}
\label{eq:sticking_contact}
\forall c\in\{c|\Omega_c(t)\neq 0 \}\,, r_{c}(t+1) = r_{c}(t).
\end{align}
Note that this constraint also implies that the end-effector cannot change its contact surface in one step; it has to break the current contact before switching to a new surface. 

\subsubsection{Coulomb friction}
We assume all contact forces satisfy the Coulomb friction model. Hence, the sticking contact should stay inside the linearized friction cone of the given contact surface. The sliding contact (only environment contacts can slide) should satisfy $f_e^{\parallel}(t) = -\mu_e\norm{f_e^{\bot}(t)}\hat{\dot{r}}_e^{\parallel}(t)$, where $f_e^{\parallel}(t)$ is the tangential force, $f_e^{\bot}(t)$ the normal force, and $\hat{\dot{r}}_e^{\parallel}(t)$ the unit direction of the contact point velocity projected onto the contact surface, which can be obtained from the given object motion. For notational simplicity, we denote these constraints by the respective feasible set, hence
\begin{align}
\label{eq:coulomb_friction}
f_c(t) \in \mathbb{F}_c(t)\,, f_e(t) \in \mathbb{F}_e(t)
\end{align}

\subsubsection{Cost function}
Finally, we minimize a quadratic objective function that avoids applying large forces at the boundary of the contact surface
\begin{align}
    J = \sum_{t=1}^{T}\sum_{c=1}^{N_c}\norm{f_c(t)}^2 +\norm{r_c(t)}^2
\end{align}

\subsubsection{Biconvex Decomposition}
The optimization problem described above has an interesting feature that the only non-convex constraint~\eqref{eq:euler_constr} due to the cross product $r_c(t) \times f_c(t)$ is in fact biconvex. That is, when $r_c(t)$ is fixed, this constraint is convex with respect to $f_c(t)$; when $f_c(t)$ is fixed, this constraint is convex with respect to $r_c(t)$. Note that in both cases, the environment contact location $r_e(t)$ is known and thus the cross product $r_e(t) \times f_e(t)$ does not pose any non-convexity. In fact, when we group the decision variables into two sets $x = [r_c(t), \alpha_c(t)]_{t=1}^{T}$ and $z = [f_c(t), f_e(t)]_{t=1}^{T}$, we can re-write the original problem into the standard \ac{ADMM} form with the constraint 
\begin{align*}
    G(x, z) &= \sum_{c=1}^{N_c} r_c(t) \times f_c(t) +\sum_{e=N_c + 1}^{N_c + N_e(t)} r_e(t) \times f_e(t) - \tau_{\text{des}}(t)\\
    &=0\,,
\end{align*}
and the iteration
\begin{subequations}
\label{eq:admm}
\begin{align}
&x^{k+1} = && \argmin_{x}  \sum_{t=1}^{T}\sum_{c=1}^{N_c}\norm{r_c(t)}^2
    +\frac{\rho}{2}\norm{G(x, z^k) + y^k}
\label{eq:location_problem}\\
& && \text{ s.t. } \eqref{eq:contact_location}, \eqref{eq:sticking_contact}\notag
\\
&z^{k+1} = && \argmin_{z} \sum_{t=1}^{T}\sum_{c=1}^{N_c}\norm{f_c(t)}^2 +
    \frac{\rho}{2}\norm{G(x^{k+1}, z)+y^k}
\label{eq:force_problem}\\
& && \text{ s.t. }\eqref{eq:newton_constr}, \eqref{eq:force_complementarity},\eqref{eq:coulomb_friction}\notag
\\
&y^{k+1}= && y^k + G(x^{k+1}, z^{k+1})\,,
\end{align}
\end{subequations}
where $y$ is the scaled dual variable and $\rho > 0$ is a penalty parameter that is tuned to $2\times10^6$ in the experiments. The \ac{ADMM} iteration initializes with $\alpha_{c,i}^0(t) = 1/N_{v, \Omega_c(t)}$, $r_c^0(t) = \sum_{i=1}^{N_{v, \Omega_c(t)}}\alpha_{c,i}^0(t)v_{\Omega_c(t), i}$, $f_c^0(t) = 0$, $f_e^0(t)=0$ and $y^0 = 0$. Note that both the optimization problems~\eqref{eq:location_problem} and~\eqref{eq:force_problem} are just \acp{QP}~\cite{boyd2004convex} which are simple to solve. Throughout our experiments, we only run one \ac{ADMM} iteration as we observe satisfactory convergence in this setting. Hence, solving this non-convex optimization problem only requires solving two \acp{QP}.

\subsection{Discrete Contact Planning via MCTS}
\begin{figure}
    \centering
    \includegraphics[width=.45\textwidth]{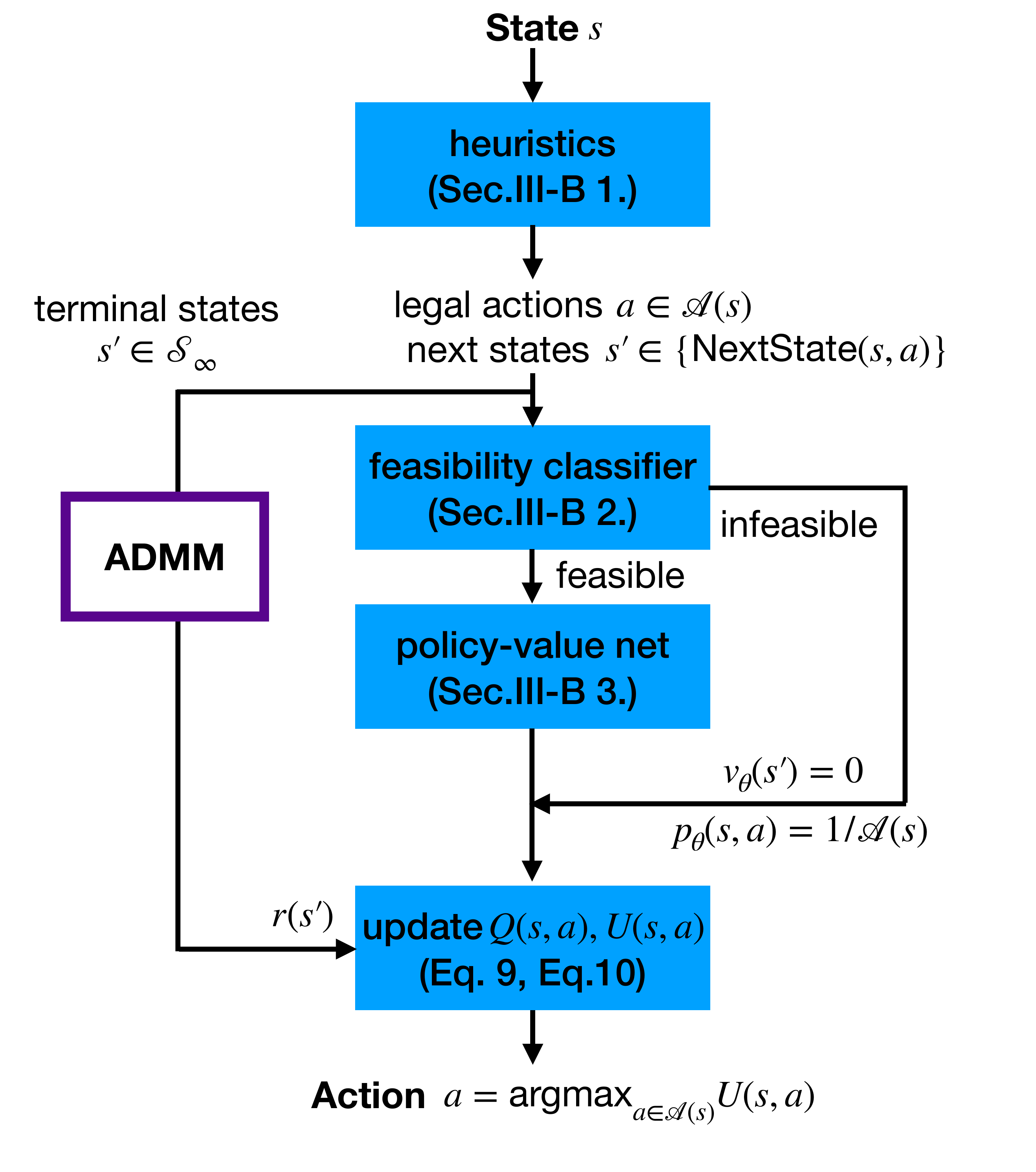}
    \caption{Action selection in the MCTS search process. The actions are evaluated by both domain-specific heuristics and trained neural networks to enable efficient search. As the search reaches a terminal state, the reward is computed by ADMM.   For readability, recursions are omitted.}
    \label{fig:mcts}
    \vspace{-5mm}
\end{figure}

Now that we have shown the contact locations and forces can be found efficiently if the contact surfaces are known, we focus on the missing piece: finding these contact surfaces. To do this, we adopt \ac{MCTS} that was behind many recent advances in reinforcement learning for game-play~\cite{silver2017mastering, schrittwieser2020mastering}. To solve the contact sequence discovery problem via \ac{MCTS}, we first model it as a \ac{MDP} with states $s\in \mathcal{S}$ and actions $a\in \mathcal{A}(s)$, where $\mathcal{S}$ is the set of states and $\mathcal{A}(s)$ is the set of legal actions at the state $s$. At time step $k$, the action $a_{k} = [\Omega_1(k),\dots,\Omega_{N_c}(k)]$ represents which object surface each end-effector needs to be in contact with (or not) and the state $s_k$ contains the current desired object pose $q(k)$ and the history of the object surfaces that were in contact with each end-effector $[\Omega_1(t),\dots,\Omega_{N_c}(t)]_{t=1}^{k}$. Note that the state transition $s_{k+1}=\textsc{NextState}(s_k, a_k)$ in this \ac{MDP} is deterministic as the desired pose is known a priori and thus fixed. With this formulation, we modify the \ac{MCTS} algorithm by utilizing both domain-specific heuristics and neural networks trained from past experience to improve search efficiency. As the search reaches the end, an optimization problem is constructed and solved by \ac{ADMM} to provide a terminal reward. Fig.~\ref{fig:mcts} previews these modifications that will be elaborated in the remainder of this section.

Let us first recall the standard learning-based \ac{MCTS} algorithm as summarized in Algorithm~\ref{alg:mcts}. It constructs a search tree $\mathcal{T} = (\mathcal{V}, \mathcal{E})$ where the set of vertices $\mathcal{V}$ contains the visited states and the set of edges $\mathcal{E}$ contains the visited transitions $(s\overset{a}{\rightarrow}s')$. For each transition, it maintains the state-action value $Q(s, a)$ that estimates future rewards to be accumulated and the number of occurrences $N(s, a)$ of this state-action pair during the search. To update the state-action value $Q(s, a)$, learning-based \ac{MCTS} uses a policy-value network (parameterized by ${\theta}$) to provide a value estimate $v_{\theta}(s')$ for the possible next states $s'\in \{\textsc{NextState}(s, a)|a \in \mathcal{A}(s)\}$
\begin{align}
    Q(s, a) \gets \frac{N(s,a)Q(s,a)+v_{\theta}(s')}{N(s,a)+1}\,.
\end{align}
The same network also outputs an action probability $p_{\theta}(s, a)$ to calculate a heuristic score for the state-action pair
\begin{align}
    U(s, a) = Q(s, a) + \gamma p_{\theta}(s, a)\frac{\sqrt{\sum_{b}N(s, b)}}{1+N(s, a)}\,,
\end{align}
where $\gamma>0$ is a hyper-parameter controlling the degree
of exploration; in our experiments, it is manually tuned to $0.1$. This score is used by \ac{MCTS} to select promising actions $a = \argmax_{a\in \mathcal{A}(s)}U(s, a)$ while balancing exploration with exploitation. Once the search reaches a terminal state $s \in \mathcal{S}_{\infty}$, the contact surfaces found by \ac{MCTS} are used to construct the optimization problem described in Sec.~\ref{sec:cont_opt}. Its solution will then be evaluated to return a reward $r$ to update the state-action value and guide future search.

To calculate the reward, we integrate the solution to obtain the final object pose $\hat{q}(T) = [\hat{p}(T), \hat{R}(T)] $ with the semi-implicit Euler method. We then compare it with the desired final pose $q(T)=[p(T), R(T)] $ to compute a weighted distance
\begin{align}
    D(q, \hat{q}) = \norm{p - \hat{p}} + \beta \norm{\log (\hat{R}^{\mathsf{T}}R)}\,,
\end{align}
where $\beta > 0$ scales the angular distance; in the experiments, it is set to $0.1$. Note that this distance does not always equal zero as we terminate \ac{ADMM} only one iteration. The weighted distance within a threshold $D \leq D_{\text{th}}$ is then normalized to $[0, 1]$ to obtain the reward.
In our experiments, the threshold $D_{\text{th}}$ is set to $0.03$ which allows a maximal object position error of $\SI{3}{\centi\meter}$ or an orientation error of $\SI{0.3}{\radian}$. Note that we set a relatively high threshold in order to collect diverse training data; after training, the \ac{MCTS} almost always discovers solutions with near zero error as will be shown in the experiments. If the contact sequence found by \ac{MCTS} does not lead to a dynamically feasible solution or has a higher error above the threshold, the reward will be set to zero. Hence, the $\textsc{Evaluate}(s)$ function in \ac{MCTS} entails the \ac{ADMM} iteration~\eqref{eq:admm} and computing the reward.
\begin{algorithm}
\footnotesize
\caption{Learning-Based Monte Carlo Tree Search}\label{alg:mcts}
\begin{algorithmic}[1]
\Procedure{\ac{MCTS}}{$s;\theta$}
\If{$s \in \mathcal{S_{\infty}}$}
    \State $r \gets$ \Call{evaluate}{$s$}
    \State \textbf{return} $r$ 
\ElsIf{$s \notin \mathcal{V}$}
    \State $\mathcal{V} \gets \mathcal{V} \cup \{s\}$
    \For{$ a \in \mathcal{A}(s)$}
        \State $Q(s, a) \gets 0$
        \State $N(s, a) \gets 0$
    \EndFor
    \State \textbf{return} $v_{\theta}(s)$
\Else
    \State $a \gets \argmax_{a\in \mathcal{A}(s)}U(s, a)$
    \State $s' \gets \Call{nextState}{s, a}$
    \State $v \gets \Call{\ac{MCTS}}{s';\theta}$
    \State $Q(s, a) \gets \frac{N(s,a)Q(s,a)+v}{N(s,a)+1}$
    \State $N(s,a) \gets N(s,a) + 1$
    \State $\mathcal{E} \gets \mathcal{E} \cup \{(s\overset{a}{\rightarrow}s')\}$
    \State \textbf{return} $v$
\EndIf
\EndProcedure
\end{algorithmic}
\end{algorithm}

At inference
time, the procedure $\text{MCTS}(s;\theta)$ is repeated until the first feasible solution is found; at
training time, we let it discover multiple solutions
and collect both feasible and infeasible contact sequences to
construct a diverse training set $\mathcal{D}=\{(\bar{v}(s), \bar{p}(s,a)) | s \in \mathcal{V}\}$ for all visited states, where $\bar{p}(s, a)=N(s, a)/\sum_b N(s,b)$ is the empirical action probability and $\bar{v}(s)=\sum_{a \in \mathcal{A}(s)} \bar{p}(s, a)Q(s, a)$ is the expected state value. The network is then updated using the sum of a mean-square loss $l_v$ for the value head and a cross-entropy loss $l_{p}$ for the policy head
\begin{align}
    l(\mathcal{D}) = \frac{1}{|\mathcal{D}|}\sum_{s\in \mathcal{V}}\big(l_v(s) + l_{p}(s)\big)\,,
\end{align}
where 
\begin{subequations}
\begin{align}
l_v(s) &= \big(v_{\theta}(s) - \bar{v}(s)\big)^2\\
l_{p}(s) &= -\sum_{a \in \mathcal{A}(s)}\bar{p}(s,a)\log p_{\theta}(s, a)\,.
\end{align}
\end{subequations}

\subsubsection{Assumptions and Heuristics}
\label{sec:heuristics}
One major advantage of using \ac{MCTS} over \ac{MIQP} is that it is straightforward to incorporate domain-specific assumptions and heuristics to reduce the search space. In this work, we apply the following assumptions and heuristics:
\begin{itemize}
    \item \textbf{Contact surface:} Each contact surface can only be touched by at most one end-effector and each end-effector can touch at most one contact surface.
    
    \item \textbf{Persistent contact:} While the downstream continuous optimization problem may have a small discretization step, for example $\Delta t = \SI{0.1}{\second}$, most manipulation tasks do not require decisions of contact switch  at such a high resolution. Thus, we assume that an end-effector must remain on the same surface for $d$ time steps, which means a trajectory of length $T$ can admit at most $T/d - 1$ contact switches.

    \item \textbf{Kinematic feasibility:} For each end-effector $c$, a contact surface will only be considered if inverse kinematics can find a robot configuration that reaches the center of this surface within an error threshold of $\SI{2}{\centi\meter}$ and does not result in any undesired collision (e.g. between non-end-effector links and the object). Note that this cannot be imposed in a \ac{MIQP} formulation as it introduces nonlinear constraints.
    
    \item \textbf{Contact switch:} We allow at most one end-effector to make or break the contact at each time step. Moreover, the end-effector can only break the contact if the desired object velocity and acceleration is zero.
\end{itemize}

\subsubsection{Feasibility classifier}
One key difference between our task and generic game-play is that our dataset is highly imbalanced---many contact sequences explored by the \ac{MCTS} are dynamically infeasible, resulting in zero reward. Directly training on such a dataset leads to underestimation of the value function. Instead, we first train on the dataset $\mathcal{D}$ a binary classifier $C_{\phi}(s)$ with logistic regression where dynamically feasible samples are given more weights. At inference time, a state is only fed into the policy-value network if the classifier labels it as dynamically feasible; otherwise, we output zero value $v_{\theta}(s)=0$ and uniformly distributed action probability $p_{\theta}(s,a)=1/|\mathcal{A}(s)|$. This feasibility classifier screens out dynamically infeasible contact sequences before the \ac{MCTS} completes the search, thus greatly improves search efficiency.

\subsubsection{Goal-conditioned policy-value network}
\begin{figure}
    \centering
    \includegraphics[width=.45\textwidth]{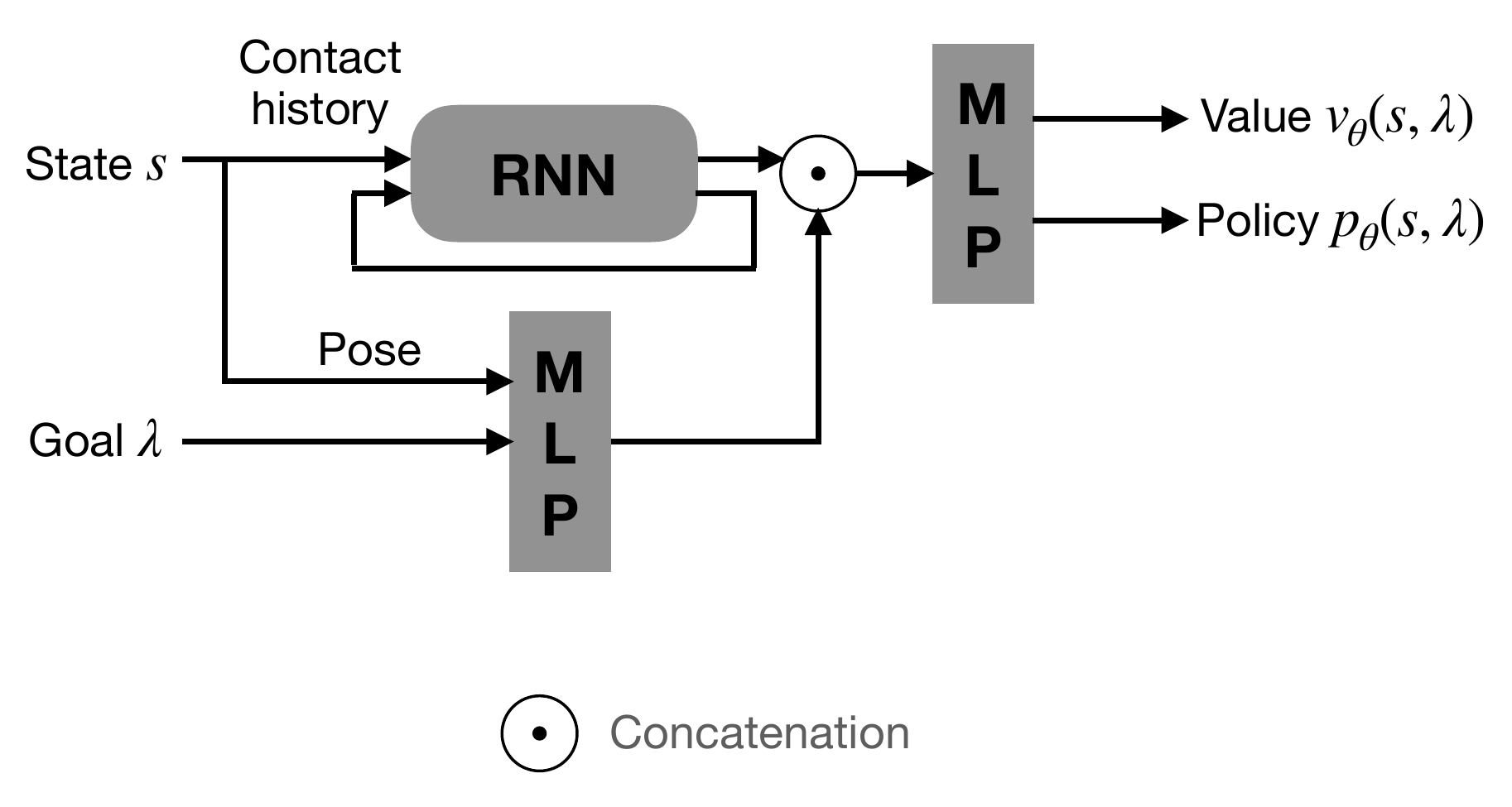}
    \caption{Schematic diagram of the policy-value network architecture. Activation functions and regularization layers such as Dropout and BatchNorm are omitted.}
    \label{fig:nn}
    \vspace{-5mm}
\end{figure}

Note that each \ac{MCTS} instance only searches for the contact sequence for a given object motion $\xi$, thus the rewards are motion-specific. If we were to learn from the data collected for this object motion only, it is unlikely that the network would generalize to other motions. Thus, we generate multiple object motions in the training phase and additionally input an intermediate goal variable to the network. It is defined as the difference between the current desired pose and the one $h$ steps in the future $\lambda(t) = q(t+h)\ominus q(t)$, where $\ominus$ denotes subtraction in $\mathrm{SE}(3)$. Fig.~\ref{fig:nn} depicts the policy-value network architecture: it takes as inputs the state $s$ and the goal $\lambda$, and outputs the goal-conditioned value $v_{\theta}(s, \lambda)$ and action probabilities $p_{\theta}(s, \lambda)$. Since the sequence of contact surfaces has varying lengths, we use a \ac{RNN} to encode this information and concatenate it with the pose and the goal processed by a \ac{MLP}. Due to the usage of the feasibility classifier mentioned above, we only train our policy-value network on a subset $\mathcal{D}_+ \subseteq \mathcal{D}$ with positive samples $\mathcal{V}_+=\{s\in \mathcal{V}|\bar{v}(s) > 0\}$ to avoid underestimating the value function.
\section{Experiments}
We conduct experiments in simulation and on real hardware to show that our method \begin{enumerate*}
    \item is capable of finding high quality dynamically feasible solutions much faster than a \ac{MIQP} baseline.
    \item scales to long-horizon tasks even when trained only on data collected from shorter-horizon tasks. 
\end{enumerate*}

\subsection{Experiment Setup}

\subsubsection{Tasks}
\label{sec:task}
Throughout all experiments, we consider a manipulator composed of two modular robot fingers similar to the ones used in~\cite{wuthrich2020trifinger} and a $\SI{10}{\centi\meter} \times \SI{10}{\centi\meter} \times \SI{10}{\centi\meter}$ cube with mass $m=\SI{0.5}{\kilo\gram}$ on an infinitely large plane. The cube and the plane have the same friction coefficient $\mu = \mu_e = 0.8$. We consider one contact surface for each face of the cube except for the bottom one; each contact surface is a $\SI{8}{cm} \times \SI{8}{cm}$ square to avoid contact locations at the corner. The object motion trajectory is generated with spline interpolation in $\mathrm{SE}(3)$ between the initial object pose and a desired pose parameterized as the following primitives:
\begin{enumerate*}
    \item \ac{S} on the $xy$-plane by \SIrange{-10}{10}{\centi\meter}
    \item \ac{SC} on the $xy$-plane by \SIrange{-5}{5}{\centi\meter} with a rotation about the $z$-axis by \SIrange{-45}{45}{\degree}
    \item \ac{R} about the $z$-axis by \SIrange{-90}{90}{\degree}
    \item \ac{L} along the $z$-axis by \SIrange{0}{10}{\centi\meter}, and
    \item \ac{P} about the $y$-axis by \SIrange{0}{45}{\degree}.
\end{enumerate*}
The $xy$-axes span the plane that the object is placed on and the $z$-axis points to the opposite gravity direction. Finally, the initial object position is randomly sampled on the $xy$-plane within a $[-\SI{5}{\centi\meter}, \SI{5}{\centi\meter}]^2$ area centered at the origin and the initial orientation about the $z$-axis by \SIrange{-90}{90}{\degree}. 

\subsubsection{Baselines}
We compare our method (\textbf{\ac{MCTS}}) with two baselines: 
\begin{itemize}
    \item the \textbf{\ac{MIQP}} baseline is implemented following~\cite{aceituno2020global}. We did not use the authors' \href{https://github.com/baceituno/QuasiDynamics}{open-source implementation} as it was only implemented for 2D objects. But the same formulation can be directly extended to 3D. To facilitate a fair comparison, we also implemented all heuristics described in Sec~\ref{sec:heuristics} except the kinematic feasibility check. For the McCormick envelope relaxation of the cross product, we partition the contact location into $4$ intervals and the contact force into $2$ intervals. In all experiments, we terminate the \ac{MIQP} solver at the first feasible solution instead of waiting for the global optimum which may take extremely long time. In addition, we implement the constraint~\eqref{eq:newton_constr} as a penalty term in the cost function. This accelerates the \ac{MIQP} solver significantly from our observation during the experiments. We note that our implementation has comparable computation time as reported in~\cite{aceituno2020global}.
    \item the \textbf{MCTS$^U$} baseline represents an untrained model and constantly outputs zero values $v_{\theta}(s) = 0$ and uniform action probability $p_{\theta}(s, a) = 1/|\mathcal{A}(s)|$.
\end{itemize}

\subsubsection{Software and hardware}
We conduct all experiments on a single GeForce RTX 2080 TI GPU and an Intel Xeon CPU at \SI{3.7}{\giga\hertz} using Python and PyTorch. We model and solve the \acp{QP} with CVXPY~\cite{diamond2016cvxpy} and OSQP~\cite{osqp} and use Gurobi~\cite{gurobi} for \ac{MIQP}. All source code including the baseline can be found at \url{https://hzhu.io/contact-mcts}.

\subsection{Evaluation metrics}
We examine two metrics to evaluate the effectiveness and efficiency of our method:
\begin{enumerate*}
    \item the \textbf{force and torque error} between the desired and the solution. The error is averaged over the horizon $T$ and scaled by the object mass and inertia respectively.
    The smaller this error is, the better the solution tracks the desired object motion.
    \item The \textbf{computation time} needed to find the first dynamically feasible solution.
\end{enumerate*}

\subsection{Training procedure} We generate $300$ object motion trajectories, each comprising two primitives with randomly sampled parameters. In particular, $200$ trajectories are composed of two \ac{SC} primitives; $50$ trajectories of one \ac{SC} and one \ac{L}; $50$ trajectories of one \ac{SC} and one \ac{P}. For the $i$-th trajectory, we let an untrained \ac{MCTS} run until it evaluates $200$ candidate contact sequences; then we construct the dataset $\mathcal{D} = \mathcal{D} \cup \{(\bar{v}(s), \bar{p}(s)) | s \in \mathcal{V}_i\}$ where $\mathcal{V}_i$ contains all the states the \ac{MCTS} visited for the $i$-th trajectory. The policy-value network and the value classifier are then trained via Adam~\cite{kingma2014adam} for $300$ epochs.

\subsection{Single motion primitives}
\label{sec:primitive}
In this experiment, we consider object motion trajectories that consist of one single primitive. Each primitive has a desired pose uniformly randomly sampled from its respective parameter range described in Sec.~\ref{sec:task}. Each trajectory consists of $T=10$ time steps with step size $\Delta t = \SI{0.1}{\second}$; each contact persists as well at least \SI{0.1}{\second}, hence a trajectory can admit at most $9$ contact switches.  We run $50$ trials for each primitive to collect the performance statistics. 

Table~\ref{tab:primitive} shows that our method is capable of finding dynamically feasible solutions consistently faster than the \ac{MIQP} baseline thanks to the \ac{MCTS} formulation. Especially for primitives that require non-zero torques (\ac{R}, \ac{SC}, \ac{P}), the \ac{MIQP} baseline is not only an order of magnitude slower, but also produces solutions with large errors. We note that the force error can be reduced by letting the \ac{MIQP} solver explore more feasible solutions, while the torque error remains high nonetheless. This might be due to its usage of the McCormick envelopes to approximate cross products, which not only introduces approximation error but also adds additional discrete variables. In contrast, thanks to the \ac{ADMM} formulation, our method solves the original problem instead of a relaxed one and has thus near-zero average force and torque error.

We also note that while the solutions found by the \ac{MIQP} baseline are dynamically feasible, they are not necessarily kinematically feasible or collision-free since these conditions cannot be incorporated as linear constraints. While it is possible to collect multiple dynamically feasible solutions and pick the kinematically feasible one from them, it may further increase the computation time.

\begin{table}[t]
\scriptsize
\centering
\caption{Task performance for object motion primitives. Values $\leq 0.005$ are rounded to zero.}
\label{tab:primitive}
\begin{tabular}{cccccc}
\toprule
& \multirow{2}{*}{\textbf{Method}} & \multicolumn{2}{c}{\textbf{Time} [\SI{}{\second}]} & \multicolumn{2}{c}{\textbf{Error [\SI{}{\newton}, \SI[inter-unit-product =\cdot]{}{\newton\meter}]}}  \\
 & & Mean & Worst & Mean & Worst\\
\midrule
\multirow{3}{*}{\ac{S}} & MIQP & $0.65$       & $0.79$           & $0.66, \mathbf{0.00}$ & $2.90, \mathbf{0.00}$ \\
&  MCTS & $\mathbf{0.10}$ & $\mathbf{0.18}$       & $\mathbf{0.00, 0.00}$           & $\mathbf{0.00, 0.00}$      \\
&  MCTS$^U$ & $0.24$ & $1.23$       & $\mathbf{0.00}, 0.03$           & $0.06, 1.14$      \\
\midrule
\multirow{3}{*}{\ac{L}}  & MIQP    & $0.25$ & $0.51$       & $6.29,  \mathbf{0.00}$           & $6.87, \mathbf{0.00}$\\
 & MCTS & $\mathbf{0.15}$ & $\mathbf{0.23}$       & $\mathbf{0.24},  0.05$           & $\mathbf{0.86}, 0.18$      \\
 & MCTS$^U$ & $0.53$ & $2.23$ & $0.53, 0.11$           & $0.88, 0.35$      \\
\midrule
\multirow{3}{*}{\ac{R}} & MIQP & $4.83$ & $29.46$       & $8.36, 16.64$ & $30.72, 45.57$        \\
 & MCTS & $\mathbf{0.12}$ & $\mathbf{0.27}$       & $\mathbf{0.00, 0.00}$           & $\mathbf{0.00, 0.00}$      \\
& MCTS$^U$ & $0.41$ & $1.22$ & $\mathbf{0.00, 0.00}$ & $\mathbf{0.00, 0.00}$      \\
\midrule
\multirow{3}{*}{\ac{SC}}  & MIQP & $2.19$ & $4.41$       & $11.73, 22.39$           & $49.61, 88.27$      \\
 & MCTS & $\mathbf{0.11}$ & $\mathbf{0.24}$       & $\mathbf{0.00, 0.00}$           & $\mathbf{0.00, 0.00}$      \\
  & MCTS$^U$ & $0.20$ & $0.81$       & $\mathbf{0.00, 0.00}$           & $0.03,0.07$      \\
  \midrule
 \multirow{3}{*}{\ac{P}}  & MIQP & $6.69$ & $50.41$       & $7.65, 15.31$           & $26.85, 53.65$      \\
 & MCTS & $\mathbf{0.15}$ & $\mathbf{0.34}$       & $\mathbf{0.01, 1.23}$           & $\mathbf{0.23, 14.01}$      \\
 & MCTS$^U$ & $0.17$ & $0.45$ & $\mathbf{0.01}, 1.46$   & $0.33, 19.55$      \\
\bottomrule
\end{tabular}
\vspace{-0.5cm}
\end{table}

\subsection{Long-horizon tasks}
In the previous experiments, we have shown the effectiveness of our method for short motion primitives. Let us now consider tasks that last a longer period of time. First, we extend the primitive to $T=30$ time steps by stipulating each contact persists for $d=3$ steps. Note that there are still at most $9$ contact switches for a single primitive. However, such extended primitives may be useful for tasks that require longer execution time but not necessarily more contact switches; for instance, sliding a cube for a long distance or rotating it very slowly. In the following experiments, we concatenate such extended primitives to form a even longer trajectory. In particular, we consider the primitive \ac{SC} as it represents typical planar repositioning/reorienting tasks. 

Table~\ref{tab:composite} reports the performance metrics for each method. A task is considered failed if no dynamically feasible solution is found within \SI{60}{\second}. We can immediately see that the trained \ac{MCTS} efficiently solves all the tasks regardless of the trajectory length, while the \ac{MIQP} baseline and the untrained \ac{MCTS} struggle in long-horizon tasks (the errors decrease because they are computed only on successful trials). Indeed, the \ac{MIQP} baseline cannot solve any tasks containing more than two primitives in \SI{60}{\second}. Interestingly, even for the task with a single extended primitive, where the number of possible contact switches does not change compared to the previous tasks in Sec.~\ref{sec:primitive}, the \ac{MIQP} baseline still need significantly more time to find a feasible solution. This could again be attributed to the McCormick envelopes as they add additional discrete variables to each time step regardless of the underlying number of contact switches.

Finally, we highlight that the \ac{MCTS} training dataset only contains object motion trajectories consisting of at most two primitives. However, Table~\ref{tab:composite} shows that our method is able to efficiently solve the longer-horizon tasks without being explicitly trained on them as our \ac{MCTS} formulation exploits the intrinsic compositionality of the task by learning a goal-conditioned policy-value network. Hence, we do not need to collect training data on large-scale, time-consuming problems as opposed to the learning-based \ac{MIP} method proposed in~\cite{nair2020solving}.

\begin{table}[t]
\scriptsize
\centering
\caption{Task performance for object motions composed of \ac{SC} primitives. Errors are computed only on successful trials. Values $\leq 0.005$ are rounded to zero.}
\label{tab:composite}
\begin{tabular}{ccccccc}
\toprule
\#& \multirow{2}{*}{\textbf{Method}} &\textbf{Success}& \multicolumn{2}{c}{\textbf{Time} [\SI{}{\second}]} & \multicolumn{2}{c}{\textbf{Error [\SI{}{\newton}, \SI[inter-unit-product =\cdot]{}{\newton\meter}]}}  \\
\textbf{\ac{SC}}&& \textbf{rate} & Mean & Worst & Mean & Worst\\
\midrule
\multirow{3}{*}{$1$} & MIQP &  $94\%$ & $10.15$       & $60.00$           & $3.40, 11.72$ & $19.93, 41.68$      \\
&  MCTS & $\mathbf{100\%}$  & $\mathbf{0.21}$ & $\mathbf{0.41}$       & $\mathbf{0.00, 0.00}$           & $\mathbf{0.00, 0.00}$      \\
&  MCTS$^U$ & $\mathbf{100\%}$ & $0.91$ & $3.67$       & $\mathbf{0.00, 0.00}$           & $0.03,0.07$      \\
\midrule
\multirow{3}{*}{$2$} & MIQP & $42\%$ & $40.93$ & $60.00$       & $4.96, 4.38$ & $16.61, 22.54$        \\
 & MCTS & $\mathbf{100\%}$ & $\mathbf{0.47}$ & $\mathbf{1.56}$       & $\mathbf{0.00,0.00}$           & $\mathbf{0.01,0.03}$      \\
& MCTS$^U$ & $\mathbf{100\%}$ & $3.08$ & $12.84$       & $\mathbf{0.00,0.00}$           & $0.03,0.07$      \\
\midrule
\multirow{3}{*}{$3$}  & MIQP & $0\%$   & $-$ & $-$ & $-$  & $-$      \\
 & MCTS & $\mathbf{100\%}$ & $\mathbf{1.35}$ & $\mathbf{8.84}$       &$\mathbf{0.00,0.00}$  & $\mathbf{0.01,0.03}$ \\
  & MCTS$^U$ & $90\%$   & $20.87$ & $60.00$       & $\mathbf{0.00, 0.00}$           & $0.01,0.04$     \\
\bottomrule
\end{tabular}
\vspace{-0.5cm}
\end{table}

\subsection{Executing the contact plan}
To validate the contact plans found by our method, we execute them in an open-loop fashion with a simple impedance controller in the PyBullet simulator~\cite{coumans2021} and on a real robot
\begin{align}
    \tau = J^{\mathsf{T}}\big(K(r_c^{\text{w}} - r^{\text{w}}) + D(\dot{r}_c^{\text{w}} - \dot{r}^{\text{w}}) + f_c^{\text{w}}\big)\,,
\end{align}
where $J$ is the end-effector Jacobian; $K, D$ are manually tuned gain matrices; $r^{\text{w}},\dot{r}^{\text{w}}$ are the position and velocity of the end-effector and $f_c^{\text{w}}, r_c^{\text{w}}, \dot{r}_c^{\text{w}}$ are the planed contact force, location and velocity, all expressed in the world frame. 

\begin{figure}[t]
\centering
  \includegraphics[width=.45\textwidth]{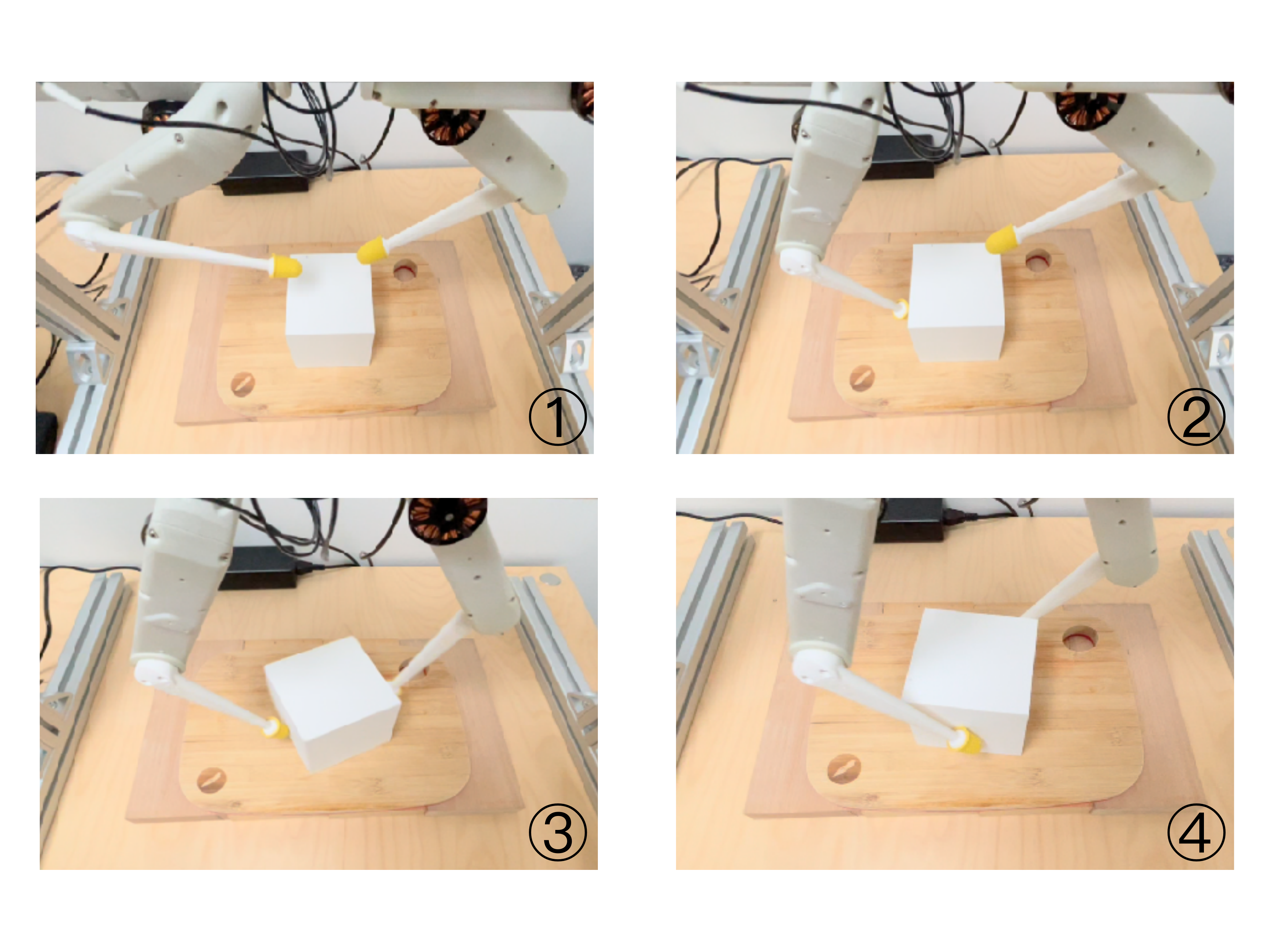}
  \caption{Exemplary execution of rotating the cube by \SI{90}{\degree}. For a video compilation of various tasks, please refer to the supplemental materials.}
  \label{fig:demos}
\end{figure}

Fig.~\ref{fig:demos} shows an example of the contact plan execution of rotating a cube by \SI{90}{\degree}. The robot is able to move the object towards the target pose if the object is placed at the desired initial position. We note that the same impedance is used for all tasks with different primitives. This would not be possible for us without applying the planed contact forces; for example, the fingers would drop the cube if they were to lift it with purely position commands and too low of a position gain. This shows the benefits and importance of planning accurate forces and torques. However, we do observe failure cases when the reality differs too much from the model, especially for the pivoting primitive \ac{P} which is sensitive to the discrepancy between the modeled and actual friction. We show this in the supplementary materials and point out the necessity of incorporating sensory feedback for online re-planning, which we leave for future work.

\section{Conclusion}
In this work, we proposed a framework that combines data-driven \ac{MCTS} and efficient non-convex optimization via \ac{ADMM} to find dynamically feasible contact forces and locations to realize a given object motion. We show that the capability of learning from data allows our method to achieve great scalability for long-horizon tasks even when the dataset only contains short-horizon data. 

The most limited aspect of our approach is that the object motion must be provided. True dexterity requires automatic discovery of object motion together with the contact plan. One potential way to achieve this is to jointly enumerate manipulator and environment contacts~\cite{cheng2022contact}. Another problem is that we represent contact surfaces as integers. While this is natural with the \ac{MCTS} formulation, it makes the learned networks object-specific. To address this issue, it might be helpful to map the integer-valued surfaces to its geometric center before feeding them into the neural networks. Finally, our approach assumes perfect knowledge of the object and the environment, which is not possible in the real world. Thus, it is necessary to explore ways of integrating perception, for example, as done in~\cite{driess2021learning}.

\bibliographystyle{IEEEtran}
\bibliography{root}

\begin{thebibliography}{10}
\providecommand{\url}[1]{#1}
\csname url@rmstyle\endcsname
\providecommand{\newblock}{\relax}
\providecommand{\bibinfo}[2]{#2}
\providecommand\BIBentrySTDinterwordspacing{\spaceskip=0pt\relax}
\providecommand\BIBentryALTinterwordstretchfactor{4}
\providecommand\BIBentryALTinterwordspacing{\spaceskip=\fontdimen2\font plus
\BIBentryALTinterwordstretchfactor\fontdimen3\font minus
  \fontdimen4\font\relax}
\providecommand\BIBforeignlanguage[2]{{%
\expandafter\ifx\csname l@#1\endcsname\relax
\typeout{** WARNING: IEEEtran.bst: No hyphenation pattern has been}%
\typeout{** loaded for the language `#1'. Using the pattern for}%
\typeout{** the default language instead.}%
\else
\language=\csname l@#1\endcsname
\fi
#2}}

\bibitem{Escande_Kheddar_2009}
A.~Escande and A.~Kheddar, ``Contact planning for acyclic motion with tasks
  constraints,'' \emph{IEEE/RSJ International Conference on Intelligent Robots
  and Systems, 2009. IROS 2009.}, p. 435–440, Oct 2009.

\bibitem{Lin_Righetti_Berenson_2020}
Y.-C. Lin, L.~Righetti, and D.~Berenson, ``Robust humanoid contact planning
  with learned zero- and one-step capturability prediction,'' \emph{IEEE
  Robotics and Automation Letters}, vol.~5, no.~2, 2020.

\bibitem{Ponton_Khadiv_Meduri_Righetti_2021}
B.~Ponton, M.~Khadiv, A.~Meduri, and L.~Righetti, ``Efficient multi-contact
  pattern generation with sequential convex approximations of the centroidal
  dynamics,'' \emph{IEEE Transactions on Robotics}, vol.~37, no.~5, p.
  1661–1679, 2021.

\bibitem{Carpentier_Tonneau_Naveau_Stasse_Mansard}
J.~Carpentier, S.~Tonneau, M.~Naveau, O.~Stasse, and N.~Mansard, ``A versatile
  and efficient pattern generator for generalized legged locomotion,'' in
  \emph{IEEE-RAS International Conference on Robotics and Automation}, 2016.

\bibitem{stewart2000implicit}
D.~Stewart and J.~C. Trinkle, ``An implicit time-stepping scheme for rigid body
  dynamics with coulomb friction,'' in \emph{IEEE International Conference on
  Robotics and Automation}, 2000, pp. 162--169.

\bibitem{posa2014direct}
M.~Posa, C.~Cantu, and R.~Tedrake, ``A direct method for trajectory
  optimization of rigid bodies through contact,'' \emph{The Int J of Robotics
  Research}, vol.~33, no.~1, pp. 69--81, 2014.

\bibitem{mordatch2012discovery}
I.~Mordatch, E.~Todorov, and Z.~Popovi{\'c}, ``Discovery of complex behaviors
  through contact-invariant optimization,'' \emph{ACM Transactions on Graphics
  (TOG)}, vol.~31, no.~4, pp. 1--8, 2012.

\bibitem{mordatch2013animating}
I.~Mordatch, J.~M. Wang, E.~Todorov, and V.~Koltun, ``Animating human lower
  limbs using contact-invariant optimization,'' \emph{ACM Transactions on
  Graphics (TOG)}, vol.~32, no.~6, pp. 1--8, 2013.

\bibitem{mordatch2012contact}
I.~Mordatch, Z.~Popovi{\'c}, and E.~Todorov, ``Contact-invariant optimization
  for hand manipulation,'' in \emph{ACM SIGGRAPH/Eurographics symposium on
  computer animation}, 2012, pp. 137--144.

\bibitem{neunert2018whole}
M.~Neunert \emph{et~al.}, ``Whole-body nonlinear model predictive control
  through contacts for quadrupeds,'' \emph{IEEE Robotics and Automation
  Letters}, vol.~3, no.~3, pp. 1458--1465, 2018.

\bibitem{pang2022global}
T.~Pang, H.~Suh, L.~Yang, and R.~Tedrake, ``Global planning for contact-rich
  manipulation via local smoothing of quasi-dynamic contact models,''
  \emph{arXiv preprint arXiv:2206.10787}, 2022.

\bibitem{toussaint2015logic}
M.~Toussaint, ``Logic-geometric programming: An optimization-based approach to
  combined task and motion planning,'' in \emph{Twenty-Fourth International
  Joint Conference on Artificial Intelligence}, 2015.

\bibitem{aceituno2020global}
B.~Aceituno-Cabezas and A.~Rodriguez, ``A global quasi-dynamic model for
  contact-trajectory optimization,'' in \emph{Robotics: Science and Systems
  (RSS)}, 2020.

\bibitem{doshi2020hybrid}
N.~Doshi, F.~R. Hogan, and A.~Rodriguez, ``Hybrid differential dynamic
  programming for planar manipulation primitives,'' in \emph{2020 IEEE
  International Conference on Robotics and Automation (ICRA)}.\hskip 1em plus
  0.5em minus 0.4em\relax IEEE, 2020, pp. 6759--6765.

\bibitem{cheng2022contact}
X.~Cheng, E.~Huang, Y.~Hou, and M.~T. Mason, ``Contact mode guided motion
  planning for quasidynamic dexterous manipulation in 3d,'' in \emph{2022
  International Conference on Robotics and Automation (ICRA)}.\hskip 1em plus
  0.5em minus 0.4em\relax IEEE, 2022, pp. 2730--2736.

\bibitem{chen2021trajectotree}
C.~Chen, P.~Culbertson, M.~Lepert, M.~Schwager, and J.~Bohg, ``Trajectotree:
  Trajectory optimization meets tree search for planning multi-contact
  dexterous manipulation,'' in \emph{IEEE/RSJ International Conference on
  Intelligent Robots and Systems}, 2021, pp. 8262--8268.

\bibitem{mccormick1976computability}
G.~P. McCormick, ``Computability of global solutions to factorable nonconvex
  programs: Part i—convex underestimating problems,'' \emph{Mathematical
  programming}, vol.~10, no.~1, pp. 147--175, 1976.

\bibitem{lazimy1982mixed}
R.~Lazimy, ``Mixed-integer quadratic programming,'' \emph{Mathematical
  Programming}, vol.~22, pp. 332--349, 1982.

\bibitem{floudas1995nonlinear}
C.~A. Floudas, \emph{Nonlinear and mixed-integer optimization: fundamentals and
  applications}.\hskip 1em plus 0.5em minus 0.4em\relax Oxford University
  Press, 1995.

\bibitem{nair2020solving}
V.~Nair, S.~Bartunov, F.~Gimeno, I.~von Glehn, P.~Lichocki, I.~Lobov,
  B.~O'Donoghue, N.~Sonnerat, C.~Tjandraatmadja, P.~Wang, \emph{et~al.},
  ``Solving mixed integer programs using neural networks,'' \emph{arXiv
  preprint arXiv:2012.13349}, 2020.

\bibitem{cauligi2021coco}
A.~Cauligi \emph{et~al.}, ``Coco: Online mixed-integer control via supervised
  learning,'' \emph{IEEE Robotics and Automation Letters}, vol.~7, no.~2, pp.
  1447--1454, 2021.

\bibitem{lin2022reduce}
X.~Lin, G.~I. Fernandez, and D.~W. Hong, ``Reduce: Reformulation of mixed
  integer programs using data from unsupervised clusters for learning efficient
  strategies,'' in \emph{2022 International Conference on Robotics and
  Automation (ICRA)}.\hskip 1em plus 0.5em minus 0.4em\relax IEEE, 2022, pp.
  4459--4465.

\bibitem{silver2017mastering}
D.~Silver \emph{et~al.}, ``Mastering the game of go without human knowledge,''
  \emph{Nature}, vol. 550, no. 7676, pp. 354--359, 2017.

\bibitem{amatucci2022monte}
L.~Amatucci, J.-H. Kim, J.~Hwangbo, and H.-W. Park, ``Monte carlo tree search
  gait planner for non-gaited legged system control,'' in \emph{2022
  International Conference on Robotics and Automation (ICRA)}.\hskip 1em plus
  0.5em minus 0.4em\relax IEEE, 2022, pp. 4701--4707.

\bibitem{gorski2007biconvex}
J.~Gorski, F.~Pfeuffer, and K.~Klamroth, ``Biconvex sets and optimization with
  biconvex functions: a survey and extensions,'' \emph{Mathematical methods of
  operations research}, vol.~66, no.~3, pp. 373--407, 2007.

\bibitem{boyd2011distributed}
S.~Boyd \emph{et~al.}, ``Distributed optimization and statistical learning via
  the alternating direction method of multipliers,'' \emph{Foundations and
  Trends{\textregistered} in Machine learning}, vol.~3, no.~1, pp. 1--122,
  2011.

\bibitem{meduri2023biconmp}
A.~Meduri, P.~Shah, J.~Viereck, M.~Khadiv, I.~Havoutis, and L.~Righetti,
  ``Biconmp: A nonlinear model predictive control framework for whole body
  motion planning,'' \emph{IEEE Transactions on Robotics}, 2023.

\bibitem{boyd2004convex}
S.~Boyd, S.~P. Boyd, and L.~Vandenberghe, \emph{Convex optimization}.\hskip 1em
  plus 0.5em minus 0.4em\relax Cambridge university press, 2004.

\bibitem{schrittwieser2020mastering}
J.~Schrittwieser, I.~Antonoglou, T.~Hubert, K.~Simonyan, L.~Sifre, S.~Schmitt,
  A.~Guez, E.~Lockhart, D.~Hassabis, T.~Graepel, \emph{et~al.}, ``Mastering
  atari, go, chess and shogi by planning with a learned model,'' \emph{Nature},
  vol. 588, no. 7839, pp. 604--609, 2020.

\bibitem{wuthrich2020trifinger}
M.~W{\"u}thrich \emph{et~al.}, ``Trifinger: An open-source robot for learning
  dexterity,'' \emph{arXiv preprint arXiv:2008.03596}, 2020.

\bibitem{diamond2016cvxpy}
S.~Diamond and S.~Boyd, ``{CVXPY}: {A} {P}ython-embedded modeling language for
  convex optimization,'' \emph{Journal of Machine Learning Research}, vol.~17,
  no.~83, pp. 1--5, 2016.

\bibitem{osqp}
B.~Stellato, G.~Banjac, P.~Goulart, A.~Bemporad, and S.~Boyd, ``{OSQP}: an
  operator splitting solver for quadratic programs,'' \emph{Mathematical
  Programming Computation}, vol.~12, no.~4, pp. 637--672, 2020.

\bibitem{gurobi}
\BIBentryALTinterwordspacing
{Gurobi Optimization, LLC}, ``{Gurobi Optimizer Reference Manual},'' 2022.
  [Online]. Available: \url{https://www.gurobi.com}
\BIBentrySTDinterwordspacing

\bibitem{kingma2014adam}
D.~P. Kingma and J.~Ba, ``Adam: A method for stochastic optimization,''
  \emph{arXiv preprint arXiv:1412.6980}, 2014.

\bibitem{coumans2021}
E.~Coumans and Y.~Bai, ``Pybullet, a python module for physics simulation for
  games, robotics and machine learning,'' \url{http://pybullet.org}.

\bibitem{driess2021learning}
D.~Driess, J.-S. Ha, and M.~Toussaint, ``Learning to solve sequential physical
  reasoning problems from a scene image,'' \emph{The Int. J of Robotics
  Research}, vol.~40, no. 12-14, pp. 1435--1466, 2021.

\end{thebibliography}
\end{document}